\documentclass[sort&compress, numafflabel]{elsarticle}

\usepackage[]{natbib}
\usepackage[breaklinks,hidelinks]{hyperref}
\usepackage{times}
\usepackage{latexsym}

\usepackage{geometry}
\usepackage{subfiles}
\usepackage{multirow}
\usepackage{array}
\usepackage{hyphenat}
\usepackage{subcaption}
\usepackage{longtable}
\usepackage{graphicx}
\usepackage{adjustbox}
\usepackage{enumitem}
\usepackage{url}

\usepackage{bm}
\usepackage{booktabs}
\usepackage{float}
\usepackage[english]{babel}
\usepackage{blindtext}
\usepackage{textcomp}
\usepackage{soul}
\usepackage{pifont}

\usepackage{longtable} 

\makeatletter
\def\ps@pprintTitle{%
 \let\@oddhead\@empty
 \let\@evenhead\@empty
 \def\@oddfoot{}%
 \let\@evenfoot\@oddfoot}
\makeatother
\usepackage{epstopdf}
\AppendGraphicsExtensions{.tif}

\usepackage{titlesec}
\titleformat{\section}
      {\normalfont\bfseries}
      {\thesection}
      {0ex}
      {\MakeUppercase}
\titleformat{\subsection}
      {\normalfont\bfseries}
      {\thesection}
      {0ex}
      {}
\titleformat{\subsubsection}
      {\normalfont}
      {\thesection}
      {0ex}
      {}

\newcommand\blfootnote[1]{%
  \begingroup
  \renewcommand\thefootnote{}\footnote{#1}%
  \addtocounter{footnote}{-1}%
  \endgroup
}

\usepackage{setspace}
\newif\ifsubfile
\subfiletrue
\newif\iftif
\tiffalse

\usepackage[utf8]{inputenc}
\usepackage[T1]{fontenc}
\usepackage{lineno}
\usepackage[numbers]{natbib}
\usepackage{subfiles}
\usepackage{graphicx}
\usepackage{float}
\usepackage{hyperref}
\usepackage{mathtools}
\usepackage{changepage}
\usepackage{caption}
\usepackage{multirow}
\usepackage{titlesec}
\usepackage{amsmath}

\usepackage{makecell}

\usepackage[math]{cellspace}
\usepackage{tikz}

\usepackage{url}

\usepackage{colortbl}
\linenumbers
\nolinenumbers

\usepackage{listings}

\title{Automated Survey Collection with LLM-based Conversational Agents}

\author[inst1]{Kurmanbek Kaiyrbekov,\textsuperscript{+}\blfootnote{\textsuperscript{+}Corresponding author: Kurmanbek Kaiyrbekov, PhD, Cyberinfrastructure and Artificial Intelligence Platforms Section, Center for Genomics and Data Science Research, National Human Genome Research Institute, National Institutes of Health, Bethesda, Maryland, USA; kurmanbek.kaiyrbekov@nih.gov}}

\ead{email@example.edu}
\author[inst2]{Nicholas J Dobbins}\ead{}
\author[inst3]{Sean D Mooney}\ead{}

\address[inst1]{Cyberinfrastructure and Artificial Intelligence Platforms Section, Center for Genomics and Data Science Research, National Human Genome Research Institute, National Institutes of Health, Bethesda, Maryland, USA}
\address[inst2]{Biomedical Informatics \& Data Science, Department of Medicine, Johns Hopkins University, Baltimore, Maryland, USA
\blfootnote{ \\ Word count: 3725} }

\begin{document}
\subfilefalse

\newpageafter{author}

\begin{abstract}

\noindent\textbf{Objective:}  Traditional phone-based surveys are among the most accessible and widely used methods to collect biomedical and healthcare data, however, they are often costly, labor intensive, and difficult to scale effectively. To overcome these limitations, we propose an end-to-end survey collection framework driven by conversational Large Language Models (LLMs).  \\

\noindent\textbf{Materials and Methods:} Our framework consists of a researcher responsible for designing the survey and recruiting participants, a conversational phone agent powered by an LLM that calls participants and administers the survey, a second LLM (GPT-4o) that analyzes the conversation transcripts generated during the surveys, and a database for storing and organizing the results. To test our framework, we recruited 8 participants consisting of 5 native and 3 non-native english speakers and administered 40 surveys. We evaluated the correctness of LLM-generated conversation transcripts, accuracy of survey responses inferred by GPT-4o and overall participant experience.   \\

\noindent\textbf{Results:} Survey responses were successfully extracted by GPT-4o from conversation transcripts with an average accuracy of 98\% despite transcripts exhibiting an average per-line word error rate of 7.7\%. While participants noted occasional errors made by the conversational LLM agent, they reported that the agent effectively conveyed the purpose of the survey, demonstrated good comprehension, and maintained an engaging interaction.   \\

\noindent\textbf{Conclusions:} Our study highlights the potential of LLM agents in conducting and analyzing phone surveys for healthcare applications. By reducing the workload on human interviewers and offering a scalable solution, this approach paves the way for real-world, end-to-end AI-powered phone survey collection systems.

\end{abstract}

\maketitle

\pagebreak

\section*{Introduction}
\label{sec:intro}

Surveys serve as a vital tool for gathering information and insight on numerous aspects of healthcare, including ensuring an accurate and unbiased assessment of clinical trials \cite{shaw2008reflux, jenkinson2005patient}. Clinical researchers and practitioners rely on surveys to screen potential participants, monitor their health over time, and ensure adherence to study protocols, making them a cornerstone of evidence-based research \cite{national2010baseline, shen1999development, lehmann2014assessing}. Healthcare facilities use surveys to measure patient satisfaction with services, enabling efforts to enhance care quality \cite{roblin2004patient, harris1997assessing}. Moreover, health surveys empower governments to identify health problems and make informed, data-driven decisions about about development of programs to improve population health. The National Center for Health Statistics (NCHS) conducts continuous surveys to assess the population's health, track progress toward national health goals \cite{cdc_implementation_evaluation_2023}, estimate health insurance coverage \cite{cohen2023health_insurance}, and evaluate care provided in physician offices and health centers \cite{santo2022characteristics}, among other objectives. These insights provide critical guidance for policymakers.   

Surveys can be conducted in various modes that differ in the manner of contacting participants, styling of questions, and the means by which the survey is distributed to participants. Self-administered surveys, such as online and mail-based formats, offer a number of advantages: they provide participants with ample time to respond thoughtfully, minimize interviewer influence, and are often cost-efficient. However, these modalities can pose challenges: mail surveys require participants to visit a post office to return completed forms, and online surveys demand basic digital literacy, which may be a barrier for older individuals. Additionally, these methods are prone to item non-response, where participants skip certain questions \cite{edwards2010questionnaires, bowling2005mode}. Face-to-face interviews, on the other hand, only require participants to speak the same language as the interviewer. They allow for real-time clarification of questions, verification of responses, and foster personal interaction, which can enhance participant engagement. However, they are resource-intensive, susceptible to interviewer bias, and geographically constrained(e.g., interviewers physically located in one locality cannot easily interview participants 10 hours away). 

Phone surveys may offer a viable middle ground. They require only telephone access (which is an increasingly universal resource) and eliminate the need for both interviewer and respondent to be in the same location. Phone surveys combine the benefits of personal engagement through audio communication with the scalability of remote administration. However, they can still be costly, time-intensive for interviewers, and subject to interviewer bias. In this study, we focus on improving the scalability of phone surveys while addressing their limitations by LLMs.

Large language models are foundational AI systems that have demonstrated exceptional performance in natural language processing tasks \cite{radford2019language, brown2020language}. Their versatility enables them to be repurposed for applications across various domains, including medicine, where they have showcased impressive reasoning capabilities by excelling in medical exams \cite{kung2023performance}. LLMs hold significant promise for transforming patient care by analyzing electronic health records and medical image data to generate accurate, interpretable diagnosis \cite{lee2025cxr}, supporting clinical decision-making through treatment recommendations aligned with established guidelines \cite{singhal2023large}, determining patients eligible for clinical trials \cite{dobbins2023leafai}, extracting relevant information from clinical notes \cite{fu2024extracting}, making health inferences from wearable sensor data \cite{kim2024health}, concisely summarizing medical evidence \cite{tang2023evaluating}, and aiding development and discovery of novel drugs \cite{liu2021ai}. Their robust instruction-following capabilities have driven the emergence of agentic AI systems, where LLMs function as specialized agents to accomplish defined tasks. These systems have been utilized to generate clinical notes from doctor-patient conversations \cite{giorgi2023wanglab}, improve performance on medical tests through multi-agent discussions with each agent serving as a specialty expert \cite{tang2023medagents}, and produce explainable clinical decisions via collaborative agent interactions \cite{hong2024argmed}. Furthermore, agentic frameworks have been proposed to streamline physicians’ administrative workloads, enhancing operational efficiency and allowing clinicians to dedicate more time to patient care \cite{gebreab2024llm, tripathi2024efficient}.

Multimodal Large Language Models (MMLLMs) have advanced significantly, expanding beyond text processing to handle and synthesize content across various modalities, including images, audio, and video \cite{zhang2024mm}. Among these, speech stands out as the most natural form of human communication. Enabling LLMs to process audio data effectively can make human-AI interactions more intuitive and seamless. Speech-processing LLMs can also be adapted into specialized agents for specific tasks, with one prominent application being telephone agents. These agents are capable of engaging in sophisticated, multilingual conversations, making them particularly well-suited for conducting surveys -- especially in the context of healthcare. In this study, we illustrate the potential of AI agents to effectively conduct health-related surveys. We propose a comprehensive AI-driven framework that not only initiates studies and administers surveys but also interprets respondents' answers and seamlessly uploads the results. We believe this innovative approach could reduce the time and effort required from human personnel, and enhance the accuracy and reliability of large-scale data collection.
%\section*{Background and Significance}
%\label{sec:background}
%No background and significance section planned.
%\subfile{sections/background}
\section*{Materials and Methods}
\label{sec:methods}

\subsection*{Survey Information}
We aimed to validate our LLM agent-based framework as effective in conducting real world surveys. To do so, we chose to use an existing survey format, the Covid Impact Survey \cite{wozniak2020covid}, rather than creating bespoke custom survey questions which may not be directly comparable to rigorously validated published studies. This survey, conducted by the National Opinion Research Center at the University of Chicago, was administered in three waves over several weeks and provided both regional and national statistics about physical health, mental health, economic security, and social dynamics. Its primary goal was to assess the immediate effects of the pandemic on the United States during a time when the federal government was preparing for the 2020 Census and unable to respond quickly. We believe that telephone-based survey collection would have been an ideal method during this period, as it could be implemented at a fast rate at scale by the federal government and aligned well with pandemic precautions such as social distancing.

From the COVID Impact Survey, we selected a subset of 33 questions to include in our study. This number was carefully chosen with comprehensiveness and respondent convenience in mind such that participants could complete one survey in under 10 minutes. While our version of the survey contained many questions related to personal, societal and economic conditions of respondents, the majority of questions were about physical and mental health. All questions required either multiple choice or numeric responses. Detailed information about our adapted survey, including the full list of questions, can be found in the \hyperref[sec:appendix]{Supplementary Appendix}.
\subsection*{System Overview}
\begin{figure}
    \centering
    \includegraphics[width=0.95\linewidth]{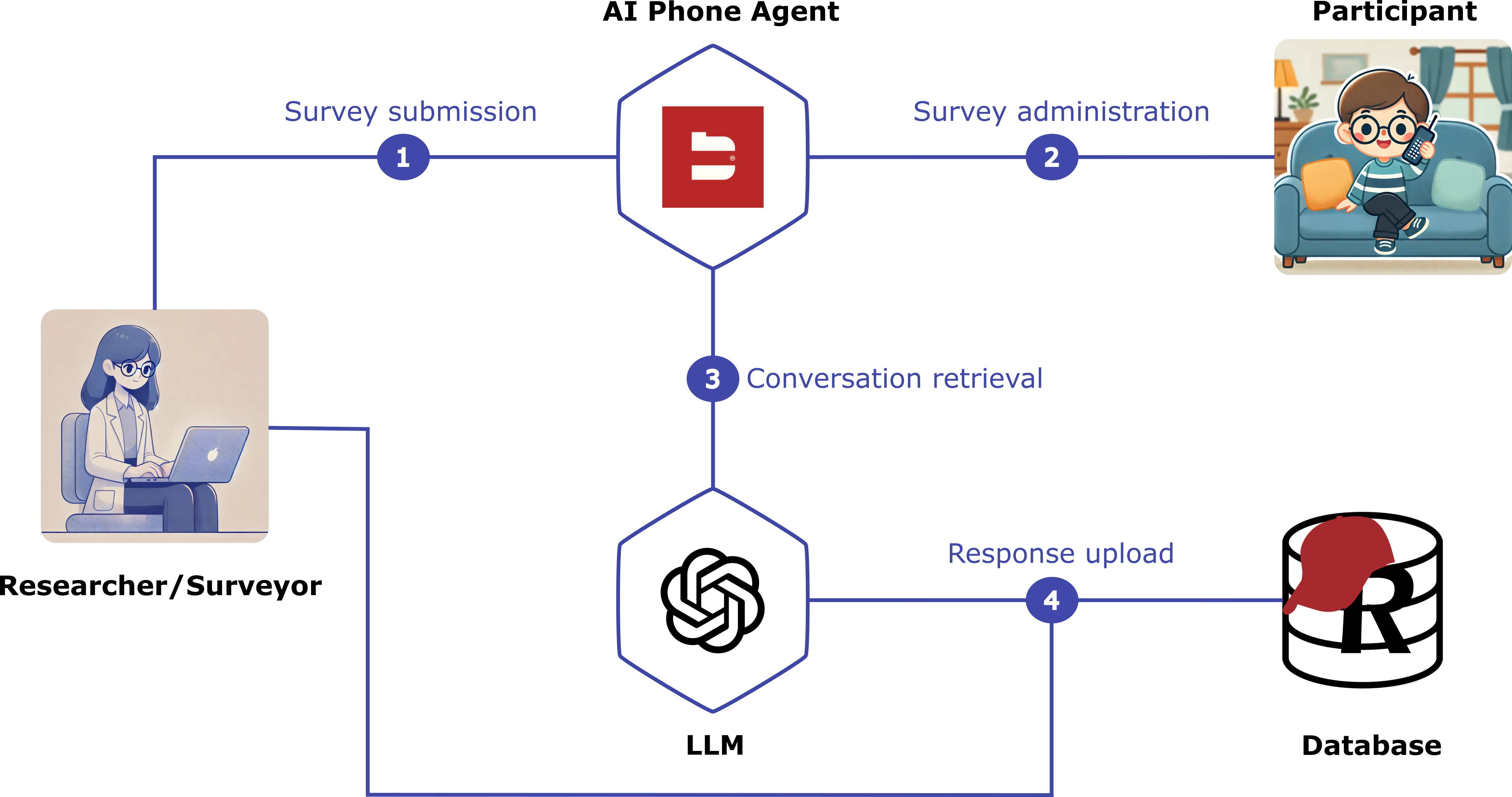}
    \caption{Overview of survey collection and analysis system. Our framework consists of a researcher that prepares a survey and writes necessary prompts for large language models, an AI-based conversable phone agent that calls participants to conducts the survey, a survey participant, a large language model that analyzes conversations to deduce answers to individual survey questions and a database for storing results.}
    \label{fig:framework}
\end{figure}
Our system comprises five main components: a researcher, an AI phone agent, the survey participants, a large language model that analyzes the surveys, and a database for storing results (Figure \ref{fig:framework}). To initiate the process, we prepare the survey and instantiate REDCap -- a secure web application designed for building and managing online surveys and databases. Then we collect participants' phone numbers, and create an instruction prompt for the AI-based phone agent. The instruction prompt is then provided to a conversable LLM-based agent developed by the company BlandAI, which calls participants, conducts the survey, and stores conversation transcripts and audio recordings. We then retrieve the transcripts and prompt GPT-4o to analyze the conversations and deduce answer to each of the questions given by the participant. Finally, we process LLM outputs and upload them into REDCap.   
\subsection*{Study Details}
\begin{figure}
    \centering
    \includegraphics[width=0.98\linewidth]{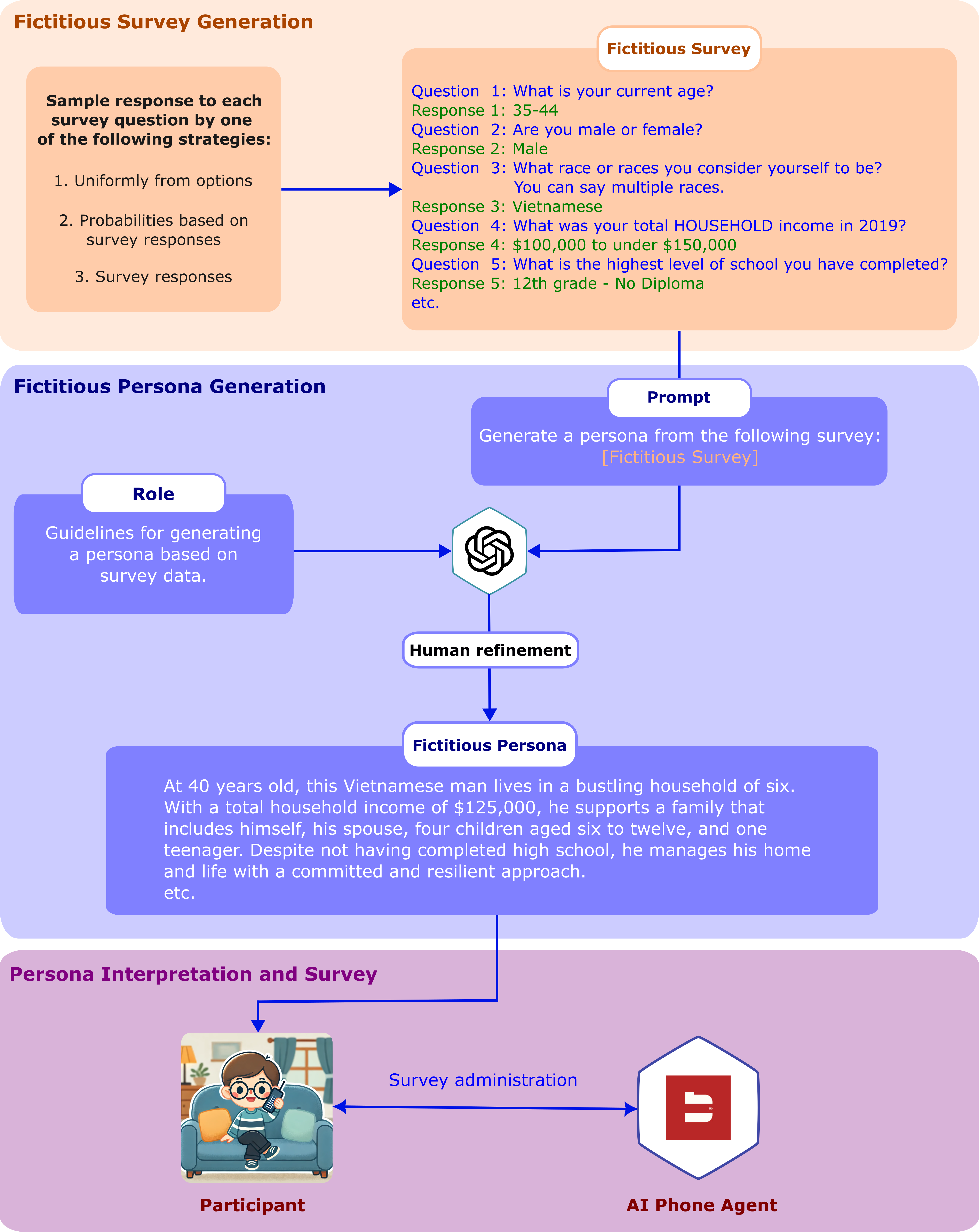}
    \caption{Representation of fictitious persona generation and survey response based on the persona. It consists of 3 main steps: 1) Fictitious survey response generation with answers to each survey question are probabilistically selected from possible options. 2) Fictitious persona generation using GPT-4o based on the fictitious survey. 3) Participant response to survey based on fictitious persona.}
    \label{fig:persona_generation_pipeline}
\end{figure}
To simulate survey collection and evaluate the quality of data gathered through our framework, we recruited N = 8 participants, primarily members of our research group to serve as survey respondents. The group included 2 female and 6 male participants. Among them, 5 were native English speakers, while 3 spoke English as a second language. This study was reviewed and approved by Institutional Review Board at NIH and granted an exemption, as it involved minimal risk, used publicly available data and did not involve personally identifiable information. Written informed consent was obtained from all participants, who were informed of the voluntary nature of participation and their right withdraw at any point without penalty.

Each participant completed the survey five times. Before each survey, they were provided with a description of an imaginary persona containing all the necessary details to answer the survey questions. This approach allowed us to collect multiple responses per participant while avoiding the collection of personally identifiable information.
The imaginary personas were generated using GPT-4o \cite{hurst2024gpt}, the flagship general-use LLM offered by OpenAI at the time of this study and which we determined to be reasonable for various tasks in this study. An overview of persona generation and survey administration process for our study in shown in Figure \ref{fig:persona_generation_pipeline}.

To create these personas, we first generated a fictitious conversation by probabilistically selecting responses to each survey question. Three methods were used for this selection: (1) uniform sampling from all possible answer choices, (2) sampling based on probabilities derived from real survey response distributions (e.g., frequencies from a COVID impact survey), or (3) random sampling from actual survey responses. Details regarding the exact generation procedure for each question and the rationale behind each choice can be found in the  \hyperref[app:persona_generation_description]{Supplementary Appendix}. Once the fictitious conversation was generated, GPT-4o was then prompted to produce a concise persona description from the transcript. To ensure that this description contained all the information needed for a participant to respond to the survey questions in a manner consistent with the generated conversation, we further checked and refined the personas. Participants were given five minutes to review and familiarize themselves with their assigned persona before speaking with the AI phone agent.

\subsection*{Evaluations}
Our evaluation and scoring focused on three key aspects:

\begin{enumerate}
    \item LLM phone agent's transcription of conversations with participants from audio to text.
    \item GPT-4o's performance in deduction of survey responses from conversation transcripts.
    \item LLM phone agent's usability for survey collection and participant experience.
\end{enumerate}

\subsubsection*{Audio to Text Transcription}
After the surveys were completed, two study team members listened to all 40 audio recordings. These recordings were securely stored on the Bland platform and were accessible only to the study team. Since the conversation transcripts resulting from the audio conversation were used to infer structured survey responses, the transcription accuracy is crucial. To verify accuracy, we created “correct transcriptions” by reviewing each Bland-generated transcript and correcting any mismatches with the source audio.

We then assessed alignment between the Bland-generated and correct (reference) transcriptions by calculating the Word Error Rate (WER) \cite{jelinek1998statisticalWER} for each participant’s response. WER measures the percentage of mis-transcribed words and is computed as follows:

\begin{equation}
    WER = \frac{S + D + I}{N}
    \label{eqn:wer}
\end{equation}
Where: 
\begin{itemize}
    \item $S$ is the number of substitutions (words incorrectly replaced).
    \item $D$ is the number of deletions (words that were omitted).
    \item $I$ is the number of insertions (extra words that were added but should not be there).
    \item $N$ is the total number of words in the correct transcription (reference).
\end{itemize}

We used the Python package \textit{jiwer} to calculate WER. Before computing WER, we standardized each response by removing empty strings, punctuation, and extra spaces; converting all text to lowercase; trimming leading and trailing spaces; and expanding common English contractions (e.g., “let’s” to “let us”).

\subsubsection*{Text Transcriptions to Structured Survey Responses}
Next, we evaluated how well a reasoning LLM (GPT-4o) could infer participant responses from potentially error-prone transcripts generated by Bland. We began by listening to all survey recordings and determining the correct responses to each question. For each transcript, we then prompted GPT-4o to deduce the survey answers based on the transcript text (as described in the \hyperref[app:instruction_for_analyzer_gpt4o]{Supplementary Appendix}). To increase reliability, we repeated this inference process five times per transcript and adopted a self-consistency prompting approach \cite{wang2023selfconsistency}, selecting the most common answer as the final response. The individual responses to survey questions were enforced to be in correct format and data types by the Python package \textit{pydantic}. Lastly, we measured the accuracy of GPT-4o’s chosen answers to gauge the correctness of the survey responses ultimately stored in the REDCap database.

\subsubsection*{Usability of Phone Agent and Participant Experience}
Finally, we aimed to understand participants’ perceptions of, and experiences with, AI-based phone agents and to assess the usability of these agents. We developed a post-study questionnaire consisting of ten Likert-scale items  and included an open-ended question for free-form feedback about the agent. Our questionnaire was primarily adapted from Chatbot Usability Questionnaire (CUQ) \cite{holmes2019usability} that assessed the personality, intelligence, error handling, and navigation of chatbots. We repurposed the CUQ to assess our LLM phone agent.  Further details on the questionnaire are provided in the \hyperref[app:post_study_questionnaire]{Supplementary Appendix}.

\section*{Results}
\label{sec:results}

Resulting WER scores of the generated transcripts are shown in Table \ref{tab:wer_scores}. The average WER across all participants was 7. 7\%, with the mean error rates for non-native group being slightly larger than the error rate for native speakers (9.6\% versus 6.4\%). 
\begin{table}[!h]  
\centering  
\caption{Per-line Word Error Rate (WER) scores for generated transcripts, categorized by speaker groups and aggregated by personas for each participant. }  
\begin{tabular}{lccccc|ccc}  
\toprule  
\textbf{Group} & \multicolumn{5}{c}{\textbf{Native Speakers}} & \multicolumn{3}{c}{\textbf{Non-Native Speakers}} \\  
\cmidrule(lr){2-6} \cmidrule(lr){7-9}  
\textbf{Participant}            & 1  & 2    & 3  & 4   & 5    & 6  & 7  & 8  \\  
 
\textbf{WER}  & $5.1$  & $6.7$  & $10.6$   & $4.9$  & $5.0$   & $6.0$   & $19.7$   & $3.2$   \\  
\textbf{Average WER} & \multicolumn{5}{c}{6.4} & \multicolumn{3}{c}{9.6} \\  
 
\bottomrule  
\end{tabular}  
\label{tab:wer_scores}  
\end{table}  

Although the conversation transcripts had transcriptions errors, the reasoning LLM (GPT-4o) was able to infer responses for survey questions with high accuracy as shown in Table \ref{tab:accuracy_scores}. The accuracies indicate no definitive correlation between Word Error Rate (WER) and accuracy. For example, participants 3 and 7 exhibited high WER in their transcriptions; however, the resulting accuracy of response deductions varied significantly -- accuracy was lower for participant 3 but remained high for participant 7. Furthermore, despite the higher transcription error rates observed among non-native speakers, GPT-4o achieved slightly greater inference accuracy for these respondents compared to native speakers.
\begin{table}[!h]  
\centering  
\caption{Average accuracy scores by speaker groups, calculated across all survey questions and aggregated by personas for each participant. All surveys were conducted in english.}  
\begin{tabular}{lccccc|ccc}  
\toprule  
\textbf{Group} & \multicolumn{5}{c}{\textbf{Native Speakers}} & \multicolumn{3}{c}{\textbf{Non-Native Speakers}} \\  
\cmidrule(lr){2-6} \cmidrule(lr){7-9}  
\textbf{Participant}            & 1  & 2    & 3  & 4   & 5    & 6  & 7  & 8  \\ 
\textbf{Accuracy}  & $98.8$  & $96.4$  & $95.8$   & $99.4$  & $98.2$   & $99.4$   & $97.6$   & $98.8$   \\  
\textbf{Average Accuracy} & \multicolumn{5}{c}{97.7} & \multicolumn{3}{c}{98.6} \\  
\bottomrule  
\end{tabular}  
\label{tab:accuracy_scores}  
\end{table} 

\begin{figure}
    \centering
    \includegraphics[width=0.98\linewidth]{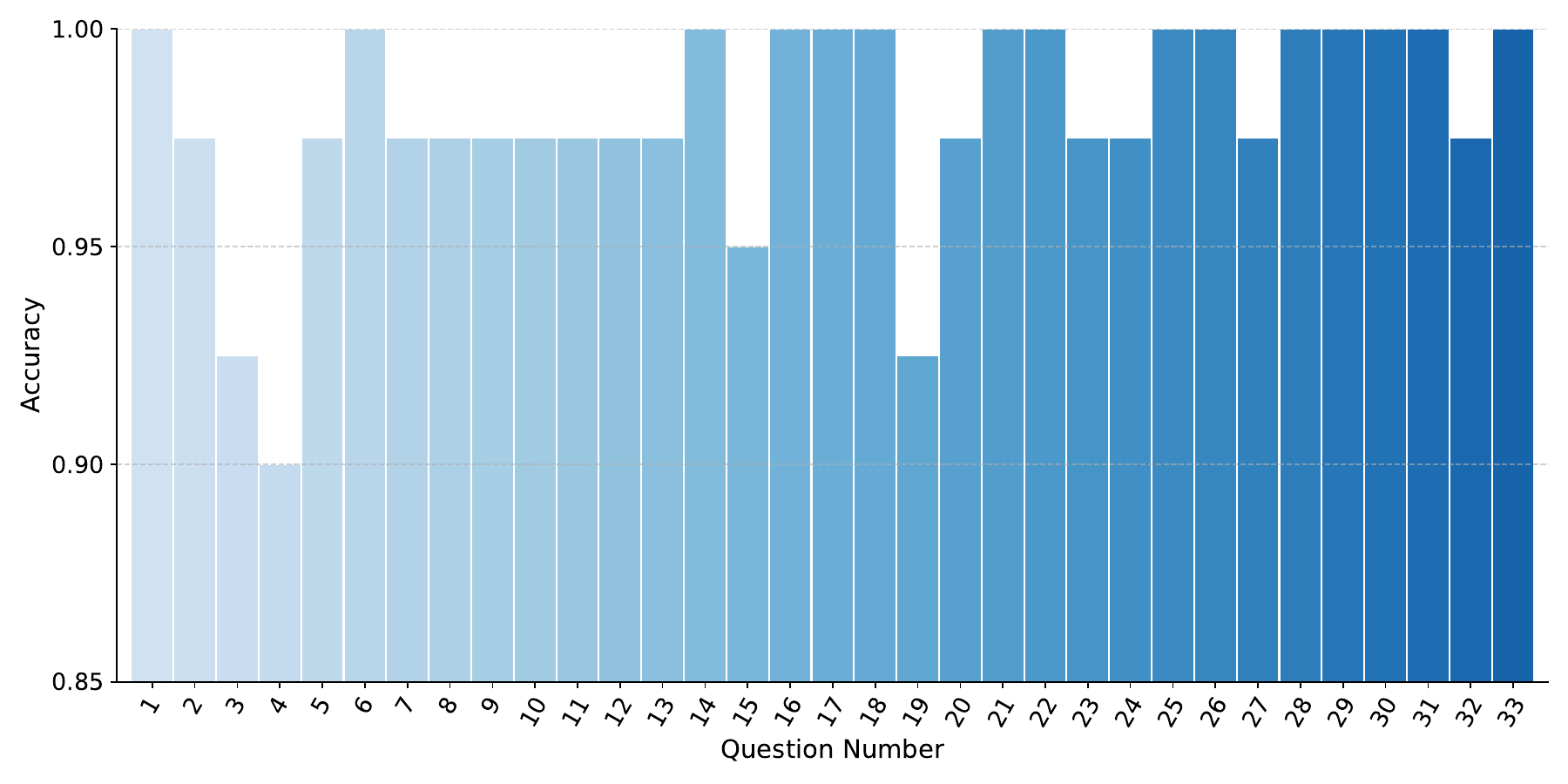}
    \caption{Average accuracy for each survey question. To compute the average accuracy, the accuracy for each question was calculated for each participant as the percentage of correct responses across five personas. The final averages were then obtained by aggregating the accuracies across all participants. }
    \label{fig:analyzer_gpt4o_accuracies}
\end{figure}

GPT-4o deduced fully correct answers for about half of the questions and word-level accuracy above 97 \% for most of the questions (see Figure \ref{fig:analyzer_gpt4o_accuracies}). The lowest score was on question 4 about the total household income. Accuracies are thus reasonably high across participants and for each of the questions.  

\begin{figure}
    \centering
    \includegraphics[width=0.98\linewidth]{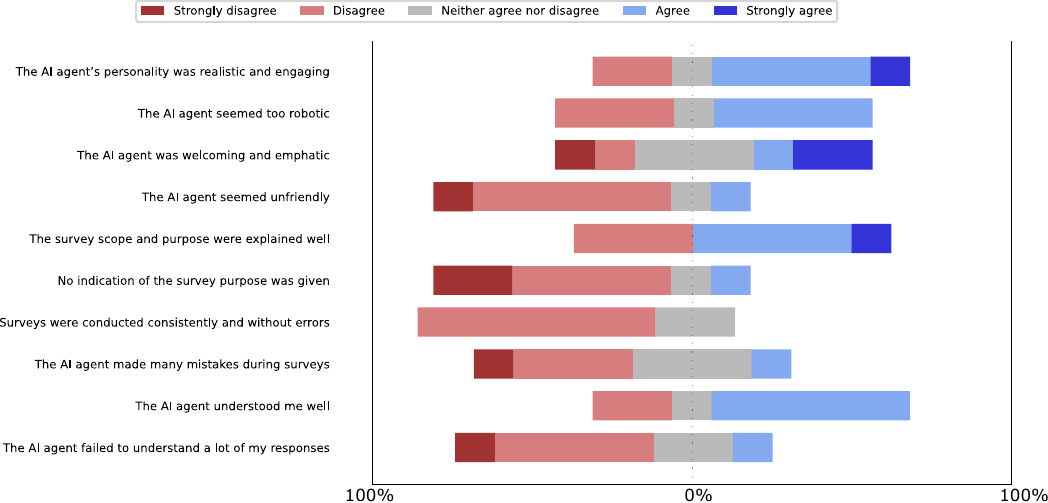}
    \caption{Results from the post-survey questionnaire, illustrating participants' agreement with various statements. Each statement is listed on the right, accompanied by horizontal bars that represent the proportion of participants selecting each response option. Bar colors correspond to specific response categories, as indicated in the legend.  }
    \label{fig:post_study_questionnaire_results}
\end{figure}

As indicated by the results of the post-survey questionnaire (Figure \ref{fig:post_study_questionnaire_results}), the majority of participants found the phone agent to be realistic and engaging, effectively explaining the purpose of the survey and demonstrating a good understanding of their responses during the surveys. Although participants acknowledged that the surveys were not entirely error-free, they did not perceive the AI agent to frequently make mistakes. Opinions were divided on whether the AI agent was empathetic, but most participants (6/8) agreed that the agent was not unfriendly. Interestingly, despite recognizing that the agent seemed realistic, many participants (4/8) also felt that it came across as slightly robotic. Additionally, in response to open ended question "What were the disadvantages of engaging with an AI agent, and what improvements could be made?", the participants mentioned that it occasionally misunderstood words for instance by confusing "male" with "mail", changed voice tone, cut off while participant was talking and sometimes spoke too fast among other things (refer to Table \ref{tab:poststudy_openended_responses} for complete reviews). We believe these feedback offer valuable insights to drive future improvements.

%Importantly, we measured this by individual words, rather than individual survey answers, as we felt overall transcription quality would be important to measure, even in cases when incorrectly transcribed words did not affect a given survey answer. We similarly measured this by accuracy.

\section*{Discussion}
\label{sec:discussion}

We found that using AI agents to conduct and analyze health-related phone surveys was highly effective, demonstrating real-world feasibility for end-to-end AI-powered survey collection. GPT-4o achieved an average response deduction accuracy of 98\% from conversation transcripts across eight participants, despite audio transcriptions exhibiting an average word error rate (WER) of 7.7\%. We believe these results could be further enhanced by leveraging more advanced large language models, such as OpenAI’s o1, and by reducing transcription errors. While participants acknowledged that the conversational AI agent was not entirely error-free, they generally felt it made few mistakes during the survey. They reported that the agent clearly explained the purpose of the survey, understood the conversation well, and was engaging. However, some felt the agent sounded slightly robotic -- a limitation we expect to improve with future iterations that focus on making the agent more human-like. Overall, while the current framework is already highly effective, continued advancements could make AI-driven survey collection virtually indistinguishable from interactions with human agents.

We envision our framework as a complementary tool to existing survey administration methods in biomedical research, such as computer-based questionnaires and face-to-face interviews, with each being used when most appropriate. By acting as an objective surveyor, the AI agent has the potential to reduce interviewer bias, a known issue where interviewer perceived expectations can influence participants to provide conforming responses. Given that current large language models demonstrate fluency in many languages, we believe this framework could also be effective for conducting surveys with international participants. In the current study, we demonstrate existing models and our framework are able to conduct surveys and gather responses reasonably well for both native- and non-native-English speakers of various backgrounds. We plan to expand this line of research by using multilingual AI agents in future. 

Another strength of this study is that our methods demonstrate notable potential for scaling and efficiency at less cost -- across significantly more participants of greater diversity -- than traditional methods. Many biomedical studies and clinical trials are challenged by the need to recruit and maintain participants, often suffering from low statistical power. Moreover, findings of traditional study methods are also challenged by poor potential for generalizability due to participants' demographics being unrepresentative of larger target populations (e.g., participants living in closer physical proximity to a medical center may be more likely to be recruited, but also more affluent and less diverse than the general population). While AI-driven studies are not a panacea for these challenges, we believe they hold great potential in vastly widening potential participation opportunities for the general population as well as simplifying operations and reducing costs for research teams. Indeed, the total costs to conduct our experiments was \$30.15 (BlandAI: \$23.92, OpenAI: \$6.23), averaging \$0.75 per survey. In future real-world scenarios, we envision that such surveys could be automatically administered to hundreds, thousands, or even millions of participants in parallel. 

Additionally, large standardized instrument databases such as Phen-X, which utilize common data elements (CDEs), offer an excellent foundation for integration. We believe many of these instruments can be seamlessly adapted for telephone-based survey administration and we intend to incorporate this capability into our framework. While our current focus is on biomedical research surveys, we see broad potential for this framework across other domains, including government census efforts, industry-specific surveys, and a wide range of academic research studies.

\subsection*{Limitations}
This project has some limitations. To avoid collecting personal information and to allow multiple surveys from a single participant, we asked participants to respond based on fictitious personas generated by GPT-4o and refined by our team. However, we observed that this approach led to interactions with the AI agent feeling less natural than if participants had been sharing their own experiences. Although we emphasized that it was not necessary to strictly adhere to the persona, and allowed participants to deviate if unsure, they occasionally paused during the call to search for information in the persona description. This behavior is atypical in natural conversations and may have impacted the flow of the interaction. Additionally, the fact that participants were aware they were speaking with an AI agent may have influenced their behavior. For example, a few participants noted that they wanted to revise an answer after responding, but refrained from doing so because the agent had already moved on to the next question. They assumed it was not possible to go back. We believe that if participants had thought they were speaking with a human, they would have felt more comfortable requesting to revise their responses.

We noticed that the phone agent sometimes wouldn't enforce the answer format, which in some cases led to ambiguous responses. Also, for some of the questions, such as questions asking about race and gender, we did not instruct the agent to list all of the response options during the survey because they had 10+ options, and we believed that the options were exhaustive. However, some of the responses were ambiguous. For example, for the question about the highest education level completed, one of the participants said, "didn't finish high school", which was ambiguous since he/she could have left school at 10th, 11th, or 12th grade, which were all separate response options. In such cases, we labeled the answer as "OTHER" and wrote in the instruction prompt to GPT-4o to also choose "OTHER" for such cases, but there were instances where GPT-4o did not choose the correct label. Similarly, another issue arose when participants gave an answer that could be construed to match multiple response options. In these cases, we felt it reasonable to choose the last uttered option as the correct response and instructed GPT-4o to do so as well. However, there were cases where GPT-4o made mistakes in this scenario. For example, for the question on whether the state of health is good, excellent, fair, or poor, the response in one of the surveys was "Good. Fair." We labeled the answer as "Fair", but GPT-4o chose "Good" as the answer. Finally, in the question about whether a person worked for pay at a job or business, even though the AI agent provided options asking if the surveyee worked for someone, was self-employed, or did not work, some of the respondents simply, "Yes." In this case, we labeled such responses as "Yes, I worked for someone." GPT-4o did not err in such scenarios. Even though such cases were very rare, we believe that the study design could have been improved if we had predicted them before the study, and potentially GPT-4o would have shown better performance.

\section*{Conclusion}
\label{sec:conclusion}

This study introduces an LLM-based framework for conducting phone surveys in healthcare and biomedical research. To illustrate the effectiveness of our framework, we conducted a study where we collected 40 surveys from 8 study participants using an AI conversational phone agent and analyzed the resulting surveys with GPT-4o. Our framework demonstrated 98\% accuracy in identifying survey responses correctly. In the future, we plan to add automatic adaptation of available survey instruments in Phen-X, and we envision that the performance of our framework will improve with better LLMs.
%TC:ignore 
\section*{Acknowledgments} 
This research was supported by the Intramural Research Program of the National Human Genome Research Institute, National Institutes of Health.

\section*{Author contributions statement}
KK is the primary developer of the framework and wrote the majority of the manuscript. NJD conducted the analysis and annotation of Bland-generated audio and transcripts, while KK handled all other analysis and annotation of the results. All authors contributed to data interpretation, manuscript revisions, and provided significant intellectual input to enhance the manuscript.

\section*{Conflict of Interests Statement}
None declared.

\bibliography{main}

\begin{thebibliography}{10}

\bibitem{shaw2008reflux}
Shaw M, Dent J, Beebe T, Junghard O, Wiklund I, Lind T, et~al.
\newblock The Reflux Disease Questionnaire: a measure for assessment of treatment response in clinical trials.
\newblock Health and quality of life outcomes. 2008;6:1-6.

\bibitem{jenkinson2005patient}
Jenkinson C, Burton JS, Cartwright J, Magee H, Hall I, Alcock C, et~al.
\newblock Patient attitudes to clinical trials: development of a questionnaire and results from asthma and cancer patients.
\newblock Health Expectations. 2005;8(3):244-52.

\bibitem{national2010baseline}
committee: NLSTRTW, Aberle DR, Adams AM, Berg CD, Clapp JD, Clingan KL, et~al.
\newblock Baseline characteristics of participants in the randomized national lung screening trial.
\newblock Journal of the National Cancer Institute. 2010;102(23):1771-9.

\bibitem{shen1999development}
Shen W, Kotsanos JG, Huster WJ, Mathias SD, Andrejasich CM, Patrick DL.
\newblock Development and validation of the diabetes quality of life clinical trial questionnaire.
\newblock Medical care. 1999;37(4):AS45-66.

\bibitem{lehmann2014assessing}
Lehmann A, Aslani P, Ahmed R, Celio J, Gauchet A, Bedouch P, et~al.
\newblock Assessing medication adherence: options to consider.
\newblock International journal of clinical pharmacy. 2014;36:55-69.

\bibitem{roblin2004patient}
Roblin DW, Becker ER, Adams EK, Howard DH, Roberts MH.
\newblock Patient satisfaction with primary care: does type of practitioner matter?
\newblock Medical care. 2004;42(6):579-90.

\bibitem{harris1997assessing}
Harris LE, Weinberger M, Tierney WM, et~al.
\newblock Assessing inner-city patients' hospital experiences: A controlled trial of telephone interviews versus mailed surveys.
\newblock Medical Care. 1997;35(1):70-6.

\bibitem{cdc_implementation_evaluation_2023}
{Centers for Disease Control and Prevention (U S )}. Implementation Evaluation: Assessing Efficiency, Effectiveness, and Impact of Public Health Programs; 2023.
\newblock Available from: \url{https://stacks.cdc.gov/view/cdc/138324}.

\bibitem{cohen2023health_insurance}
Cohen RA, Martinez ME. Health Insurance Coverage: Early Release of Estimates from the National Health Interview Survey, January–June 2023; 2023.
\newblock Available from: \url{https://stacks.cdc.gov/view/cdc/134757}.

\bibitem{santo2022characteristics}
Santo L, Schappert SM, Ashman JJ. Characteristics of Visits to Health Centers: United States, 2020; 2022.
\newblock Available from: \url{https://stacks.cdc.gov/view/cdc/117898}.

\bibitem{edwards2010questionnaires}
Edwards P.
\newblock Questionnaires in clinical trials: guidelines for optimal design and administration.
\newblock Trials. 2010;11:1-8.

\bibitem{bowling2005mode}
Bowling A.
\newblock Mode of questionnaire administration can have serious effects on data quality.
\newblock Journal of public health. 2005;27(3):281-91.

\bibitem{radford2019language}
Radford A, Wu J, Child R, Luan D, Amodei D, Sutskever I, et~al.
\newblock Language models are unsupervised multitask learners.
\newblock OpenAI blog. 2019;1(8):9.

\bibitem{brown2020language}
Brown T, Mann B, Ryder N, Subbiah M, Kaplan JD, Dhariwal P, et~al.
\newblock Language models are few-shot learners.
\newblock Advances in neural information processing systems. 2020;33:1877-901.

\bibitem{kung2023performance}
Kung TH, Cheatham M, Medenilla A, Sillos C, De~Leon L, Elepa{\~n}o C, et~al.
\newblock Performance of ChatGPT on USMLE: potential for AI-assisted medical education using large language models.
\newblock PLoS digital health. 2023;2(2):e0000198.

\bibitem{lee2025cxr}
Lee S, Youn J, Kim H, Kim M, Yoon SH.
\newblock CXR-LLAVA: a multimodal large language model for interpreting chest X-ray images.
\newblock European Radiology. 2025:1-13.

\bibitem{singhal2023large}
Singhal K, Azizi S, Tu T, Mahdavi SS, Wei J, Chung HW, et~al.
\newblock Large language models encode clinical knowledge.
\newblock Nature. 2023;620(7972):172-80.

\bibitem{dobbins2023leafai}
Dobbins NJ, Han B, Zhou W, Lan KF, Kim HN, Harrington R, et~al.
\newblock LeafAI: query generator for clinical cohort discovery rivaling a human programmer.
\newblock Journal of the American Medical Informatics Association. 2023;30(12):1954-64.

\bibitem{fu2024extracting}
Fu Y, Ramachandran GK, Dobbins NJ, Park N, Leu M, Rosenberg AR, et~al.
\newblock Extracting Social Determinants of Health from Pediatric Patient Notes Using Large Language Models: Novel Corpus and Methods.
\newblock arXiv preprint arXiv:240400826. 2024.

\bibitem{kim2024health}
Kim Y, Xu X, McDuff D, Breazeal C, Park HW.
\newblock Health-llm: Large language models for health prediction via wearable sensor data.
\newblock arXiv preprint arXiv:240106866. 2024.

\bibitem{tang2023evaluating}
Tang L, Sun Z, Idnay B, Nestor JG, Soroush A, Elias PA, et~al.
\newblock Evaluating large language models on medical evidence summarization.
\newblock NPJ digital medicine. 2023;6(1):158.

\bibitem{liu2021ai}
Liu Z, Roberts RA, Lal-Nag M, Chen X, Huang R, Tong W.
\newblock AI-based language models powering drug discovery and development.
\newblock Drug Discovery Today. 2021;26(11):2593-607.

\bibitem{giorgi2023wanglab}
Giorgi J, Toma A, Xie R, Chen SS, An KR, Zheng GX, et~al.
\newblock Wanglab at mediqa-chat 2023: Clinical note generation from doctor-patient conversations using large language models.
\newblock arXiv preprint arXiv:230502220. 2023.

\bibitem{tang2023medagents}
Tang X, Zou A, Zhang Z, Li Z, Zhao Y, Zhang X, et~al.
\newblock Medagents: Large language models as collaborators for zero-shot medical reasoning.
\newblock arXiv preprint arXiv:231110537. 2023.

\bibitem{hong2024argmed}
Hong S, Xiao L, Zhang X, Chen J.
\newblock ArgMed-Agents: Explainable Clinical Decision Reasoning with Large Language Models via Argumentation Schemes.
\newblock arXiv preprint arXiv:240306294. 2024.

\bibitem{gebreab2024llm}
Gebreab SA, Salah K, Jayaraman R, ur~Rehman MH, Ellaham S.
\newblock LLM-Based Framework for Administrative Task Automation in Healthcare.
\newblock In: 2024 12th International Symposium on Digital Forensics and Security (ISDFS). IEEE; 2024. p. 1-7.

\bibitem{tripathi2024efficient}
Tripathi S, Sukumaran R, Cook TS.
\newblock Efficient healthcare with large language models: optimizing clinical workflow and enhancing patient care.
\newblock Journal of the American Medical Informatics Association. 2024;31(6):1436-40.

\bibitem{zhang2024mm}
Zhang D, Yu Y, Dong J, Li C, Su D, Chu C, et~al.
\newblock Mm-llms: Recent advances in multimodal large language models.
\newblock arXiv preprint arXiv:240113601. 2024.

\bibitem{wozniak2020covid}
Wozniak A, Willey J, Benz J, Hart N. COVID Impact Survey: Version 1 [dataset]. Chicago, IL: National Opinion Research Center; 2020.

\bibitem{hurst2024gpt}
Hurst A, Lerer A, Goucher AP, Perelman A, Ramesh A, Clark A, et~al.
\newblock Gpt-4o system card.
\newblock arXiv preprint arXiv:241021276. 2024.

\bibitem{jelinek1998statisticalWER}
Jelinek F.
\newblock Statistical methods for speech recognition.
\newblock MIT press; 1998.

\bibitem{wang2023selfconsistency}
Wang X, Wei J, Schuurmans D, Le QV, Chi EH, Narang S, et~al.
\newblock Self-Consistency Improves Chain of Thought Reasoning in Language Models.
\newblock In: The Eleventh International Conference on Learning Representations; 2023. Available from: \url{https://openreview.net/forum?id=1PL1NIMMrw}.

\bibitem{holmes2019usability}
Holmes S, Moorhead A, Bond R, Zheng H, Coates V, McTear M.
\newblock Usability testing of a healthcare chatbot: Can we use conventional methods to assess conversational user interfaces?
\newblock In: Proceedings of the 31st European Conference on Cognitive Ergonomics; 2019. p. 207-14.

\end{thebibliography}

\clearpage

\section*{Appendix}
\label{sec:appendix}

\renewcommand{\thefigure}{S\arabic{figure}}
\setcounter{figure}{0}
\renewcommand{\thetable}{S\arabic{table}}
\setcounter{table}{0}

\section*{Adapted survey questions}
\label{app:all_survey_questions}

\begin{enumerate}
\singlespacing
\begin{spacing}{0.7}
    \item What is your current age?
    \begin{enumerate}
        \item 18-24; 
        \item 25-34; 
        \item 35-44; 
        \item 45-54; 
        \item 55-64; 
        \item 65-74; 
        \item 75+; 
        \item Under 18
    \end{enumerate}
    \item Are you male or female?
    \begin{enumerate}
        \item Male; 
        \item Female;  
    \end{enumerate}
    
    \item What race or races you consider yourself to be? You can say multiple races.
    \begin{enumerate}
        \item White; 
        \item Black or African American; 
        \item American Indian or Alaska Native; 
        \item Asian Indian; 
        \item Chinese; 
        \item Filipino; 
        \item Japanese; 
        \item Korean; 
        \item Vietnamese; 
        \item Other Asian; 
        \item Native Hawaiian; 
        \item Guamanian or Chamorro; 
        \item Samoan; 
        \item Other Pacific Islander; 
        \item Some other race; 
        \item Refused
    \end{enumerate}
    \item What was your total HOUSEHOLD income in 2019?
    \begin{enumerate}
        \item Under \$10,000
        \item \$10,000 to under \$20,000; 
        \item \$20,000 to under \$30,000; 
        \item \$30,000 to under \$40,000; 
        \item \$40,000 to under \$50,000; 
        \item \$50,000 to under \$75,000; 
        \item \$75,000 to under \$100,000; 
        \item \$100,000 to under \$150,000; 
        \item \$150,000 or more; 
        \item Don't know; 
        \item Refused
    \end{enumerate}
    \item What is the highest level of school you have completed?
    \begin{enumerate}
        \item No formal education; 
        \item 1st, 2nd, 3rd, or 4th grade; 
        \item 5th or 6th grade; 
        \item 7th or 8th grade; 
        \item 9th grade; 
        \item 10th grade; 
        \item 11th grade; 
        \item 12th grade - NO DIPLOMA; 
        \item High school graduate - high school diploma or the equivalent; 
        \item Some college, no degree; 
        \item Associate degree; 
        \item Bachelor’s degree; 
        \item Master’s degree; 
        \item Professional or Doctorate degree; 
        \item Refused
    \end{enumerate}
    \item Including yourself, how many persons currently live in your household at least 50 percent of the time? Please include any children as well as adults.
    \begin{enumerate}
        \item One person, I live by myself; 
        \item Two persons; 
        \item Three persons; 
        \item Four persons; 
        \item Five persons; 
        \item Six or more persons; 
        \item Refused
    \end{enumerate}
    
    \item How many persons currently living in your household, including yourself, are 0-1 years old?
    \begin{enumerate}
        \item  Numeric Value
        \item Don't know; 
        \item Refused;
    \end{enumerate}

    \item How many persons currently living in your household, including yourself, are 2-5 years old?
    \begin{enumerate}
        \item  Numeric Value
        \item Don't know; 
        \item Refused;
    \end{enumerate}
    \item How many persons currently living in your household, including yourself, are 6-12 years old?
    \begin{enumerate}
        \item  Numeric Value
        \item Don't know; 
        \item Refused;
    \end{enumerate}
    \item How many persons currently living in your household, including yourself, are 13-17 years old?
    \begin{enumerate}
        \item  Numeric Value
        \item Don't know; 
        \item Refused;
    \end{enumerate}
    \item How many persons currently living in your household, including yourself, are 18+ years old?
    \begin{enumerate}
        \item  Numeric Value
        \item Don't know; 
        \item Refused;
    \end{enumerate}
    \item In the past month, how often did you talk with any of your neighbors?
    \begin{enumerate}
        \item Basically every day; 
        \item A few times a week; 
        \item A few times a month; 
        \item Once a month; 
        \item Not at all; 
        \item Not sure; 
        \item Refused
    \end{enumerate}

    \item During a typical month prior to March 1, 2020, when COVID-19 began spreading in the United States, how often did you talk with any of your neighbors?
    \begin{enumerate}
        \item Basically every day; 
        \item A few times a week; 
        \item A few times a month; 
        \item Once a month; 
        \item Not at all; 
        \item Not sure; 
        \item Refused
    \end{enumerate}

    \item In the past 7 days, how often have you felt nervous, anxious, or on edge?
    \begin{enumerate}
        \item Not at all or less than 1 day; 
        \item 1-2 days; 
        \item 3-4 days; 
        \item 5-7 days; 
        \item Don't know; 
        \item Refused
    \end{enumerate}

    \item In the past 7 days, how often have you felt depressed?
    \begin{enumerate}
        \item Not at all or less than 1 day; 
        \item 1-2 days; 
        \item 3-4 days; 
        \item 5-7 days; 
        \item Don't know; 
        \item Refused
    \end{enumerate}
    \item In the past 7 days, how often have you felt lonely?
    \begin{enumerate}
        \item Not at all or less than 1 day; 
        \item 1-2 days; 
        \item 3-4 days; 
        \item 5-7 days; 
        \item Don't know; 
        \item Refused
    \end{enumerate}
    \item In the past 7 days, how often have you felt hopeless about the future?
    \begin{enumerate}
        \item Not at all or less than 1 day; 
        \item 1-2 days; 
        \item 3-4 days; 
        \item 5-7 days; 
        \item Don't know; 
        \item Refused
    \end{enumerate}
    \item In the past 7 days, how often have you had physical reactions such as sweating, trouble breathing, nausea or a pounding heart when thinking about your experience with the coronavirus pandemic?
    \begin{enumerate}
        \item Not at all or less than 1 day; 
        \item 1-2 days; 
        \item 3-4 days; 
        \item 5-7 days; 
        \item Don't know; 
        \item Refused
    \end{enumerate}

    \item In the past 7 days, did you do any work for pay at a job or business?
    \begin{enumerate}
        \item Yes, I worked for someone else for wages, salary, piece rate, commission, tips, or payments "in kind," for example, food or lodging received as payment for work performed; 
        \item Yes, I worked as self-employed in my own business, professional practice, or farm; 
        \item No, I did not work for pay last week; 
        \item Don't know; 
        \item Refused
    \end{enumerate}

    \item Would you say your health in general is excellent, very good, good, fair, or poor?
    \begin{enumerate}
        \item Excellent; 
        \item Very good; 
        \item Good; 
        \item Fair; 
        \item Poor; 
        \item Don't know; 
        \item Refused
    \end{enumerate}

    \item Has a doctor or other health care provider ever told you that you have COVID-19?
    \begin{enumerate}
        \item Yes; 
        \item No; 
        \item Not sure; 
        \item Refused
    \end{enumerate}
    \item Has a doctor or other health care provider ever told someone you live with that they have COVID-19?
    \begin{enumerate}
        \item Yes; 
        \item No; 
        \item Not sure; 
        \item Refused
    \end{enumerate}
    \item Have you experienced chills in the past 7 days, or not?
    \begin{enumerate}
        \item Yes; 
        \item No; 
        \item Not sure; 
        \item Refused
    \end{enumerate}
    \item Have you experienced runny or stuffy nose in the past 7 days, or not?\begin{enumerate}
        \item Yes; 
        \item No; 
        \item Not sure; 
        \item Refused
    \end{enumerate}
    \item Have you experienced chest congestion in the past 7 days, or not? 
    \begin{enumerate}
        \item Yes; 
        \item No; 
        \item Not sure; 
        \item Refused
    \end{enumerate}
    \item Have you experienced skin rash in the past 7 days, or not?
    \begin{enumerate}
        \item Yes; 
        \item No; 
        \item Not sure; 
        \item Refused
    \end{enumerate}
    \item Have you experienced cough in the past 7 days, or not?
    \begin{enumerate}
        \item Yes; 
        \item No; 
        \item Not sure; 
        \item Refused
    \end{enumerate}
    \item Have you experienced sore throat in the past 7 days, or not?
    \begin{enumerate}
        \item Yes; 
        \item No; 
        \item Not sure; 
        \item Refused
    \end{enumerate}
    \item Have you experienced sneezing in the past 7 days, or not?
    \begin{enumerate}
        \item Yes; 
        \item No; 
        \item Not sure; 
        \item Refused
    \end{enumerate}
    \item Have you experienced muscle or body aches in the past 7 days, or not?
    \begin{enumerate}
        \item Yes; 
        \item No; 
        \item Not sure; 
        \item Refused
    \end{enumerate}
    \item Have you experienced headaches in the past 7 days, or not?
     \begin{enumerate}
        \item Yes; 
        \item No; 
        \item Not sure; 
        \item Refused
    \end{enumerate}

    \item Can you use a thermometer to take your temperature now?
    \begin{enumerate}
        \item Yes; 
        \item No; 
        \item Don't know; 
        \item Refused
    \end{enumerate}

    \item What is your body temperature?
    \begin{enumerate}
        \item Numeric Value
        \item Refused
    \end{enumerate}
    \end{spacing}
\end{enumerate}  

An additional "OTHER" option was added to most of the questions to handle cases when answers did not fall into any category.

\clearpage

\section*{Prompt for BLAND}
\label{app:bland_prompt}

{\fontfamily{pcr}\selectfont\small \singlespacing
\begin{spacing}{1.1}
    Your name is Sarah, and you’re a surveyor for COVID-19 Household Impact Survey. The goal of the survey is to provide national and regional statistics about physical health, mental health, economic security, and social dynamics in the United States.\\
Here are the questions that you need to ask, give enough time to reply, introduce yourself, and let them know the purpose of the survey:\\  
1. What is your current age?\\  
2. Are you male or female?\\  
3. What race or races do you consider yourself to be? You can say multiple races.\\  
4. What was your total HOUSEHOLD income in 2019?\\  
5. What is the highest level of school you have completed?\\  
6. Including yourself, how many persons currently live in your household at least 50 percent of the time? Please include any children as well as adults.\\  
7. How many persons currently living in your household, including yourself, are 0-1 years old?\\  
8. How many persons currently living in your household, including yourself, are 2-5 years old?\\  
9. How many persons currently living in your household, including yourself, are 6-12 years old?\\  
10. How many persons currently living in your household, including yourself, are 13-17 years old?\\  
11. How many persons currently living in your household, including yourself, are 18+ years old?\\  
12. In the past month, how often did you talk with any of your neighbors? Response options: Basically every day, a few times a week, a few times a month, once a month, not at all, not sure?\\  
13. During a typical month prior to March 1, 2020, when COVID-19 began spreading in the United States, how often did you talk with any of your neighbors? Response options: Basically every day, a few times a week, a few times a month, once a month, not at all, not sure?\\  
14. In the past 7 days, how often have you felt nervous, anxious, or on edge? Please provide your answer in number of days.\\  
15. In the past 7 days, how often have you felt depressed? Please provide your answer in number of days.\\  
16. In the past 7 days, how often have you felt lonely? Please provide your answer in number of days.\\  
17. In the past 7 days, how often have you felt hopeless about the future? Please provide your answer in number of days.\\  
18. In the past 7 days, how often have you had physical reactions such as sweating, trouble breathing, nausea, or a pounding heart when thinking about your experience with the coronavirus pandemic?\\  
19. In the past 7 days, did you do any work for pay at a job or business? Did you work for someone, were you self-employed, or didn't work at all?\\  
20. Would you say your health in general is excellent, very good, good, fair, or poor?\\  
21. Has a doctor or other health care provider ever told you that you have COVID-19?\\  
22. Has a doctor or other health care provider ever told someone you live with that they have COVID-19?\\  
23. Have you experienced chills in the past 7 days, or not?\\  
24. Have you experienced runny or stuffy nose in the past 7 days, or not?\\  
25. Have you experienced chest congestion in the past 7 days, or not?\\  
26. Have you experienced skin rash in the past 7 days, or not?\\  
27. Have you experienced cough in the past 7 days, or not?\\  
28. Have you experienced sore throat in the past 7 days, or not?\\  
29. Have you experienced sneezing in the past 7 days, or not?\\  
30. Have you experienced muscle or body aches in the past 7 days, or not?\\  
31. Have you experienced headaches in the past 7 days, or not?\\  
32. Can you use a thermometer to take your temperature now?\\  
33. What is your body temperature?\\ 

Here is an example dialogue:\\  
You: Hello! My name is Sarah, and I’m conducting a survey to gather information on the impact of COVID-19 on various aspects of people's lives. This survey will help us understand how the pandemic has affected different communities. Your responses will be kept confidential and used for research purposes only. Could I have a few minutes of your time to answer some questions?\\  
Person: Sure!\\  
You: Thank you so much! To begin, what is your current age?\\  
Person: I’m 32 years old.\\  
You: Great, thank you. Are you male or female?\\  
Person: I’m female.\\  
You: Thank you. What race or races do you consider yourself to be? You can mention multiple races if applicable.\\  
Person: I consider myself to be Asian.\\  
You: Got it. What was your total household income in 2019?\\  
Person: Our total household income was around \$75,000.\\  
You: Thank you for sharing that. What is the highest level of school you have completed?\\  
Person: I have a Bachelor's degree.\\  
You: Including yourself, how many persons currently live in your household at least 50 percent of the time? Please include any children as well as adults.\\  
Person: There are four of us.\\  
You: And how many persons currently living in your household, including yourself, are 0-1 years old?\\  
Person: None.\\  
You: How about 2-5 years old?\\  
Person: We have one child in that age range.\\  
You: And how many are 6-12 years old?\\  
Person: None in that age range.\\  
You: How many persons currently living in your household, including yourself, are 13-17 years old?\\  
Person: None.\\  
You: Finally, how many persons currently living in your household, including yourself, are 18+ years old?\\  
Person: There are three adults.\\  
You: Thank you. In the past month, how often did you talk with any of your neighbors? Would you say basically every day, a few times a week, a few times a month, once a month, not at all, or not sure?\\  
Person: A few times a month.\\  
You: During a typical month prior to March 1, 2020, when COVID-19 began spreading in the United States, how often did you talk with any of your neighbors? Would you say basically every day, a few times a week, a few times a month, once a month, not at all, or not sure?\\  
Person: I would say a few times a week.\\  
You: In the past 7 days, how often have you felt nervous, anxious, or on edge? Please provide your answer in number of days.\\  
Person: I’ve felt that way about 3 days.\\  
You: How often have you felt depressed in the past 7 days? Please provide your answer in number of days.\\  
Person: Probably around 2 days.\\  
You: And how often have you felt lonely in the past 7 days? Please provide your answer in number of days.\\  
Person: About 4 days.\\  
You: In the past 7 days, how often have you felt hopeless about the future? Please provide your answer in number of days.\\  
Person: Maybe 1 day.\\  
You: In the past 7 days, how often have you had any physical reactions such as sweating, trouble breathing, nausea, or a pounding heart when thinking about your experience with the coronavirus pandemic?\\  
Person: I haven't experienced that.\\  
You: In the past 7 days, did you do any work for pay at a job or business? Did you work for someone, were you self-employed or didn't work at all?\\  
Person: Yes, I did work for a company.\\  
You: Would you say your health in general is excellent, very good, good, fair, or poor?\\  
Person: I would say very good.\\  
You: Has a doctor or other health care provider ever told you that you have COVID-19?\\  
Person: No, I haven't been told that.\\  
You: Has a doctor or other health care provider ever told someone you live with that they have COVID-19?\\  
Person: No, no one in my household has been told that.\\  
You: Have you experienced chills in the past 7 days, or not?\\  
Person: No, I haven't.\\  
You: Have you experienced a runny or stuffy nose in the past 7 days, or not?\\  
Person: Yes, I have.\\  
You: How about chest congestion?\\  
Person: No.\\  
You: Have you experienced a skin rash in the past 7 days, or not?\\  
Person: No.\\  
You: Have you experienced a cough in the past 7 days, or not?\\  
Person: Yes, I have.\\  
You: How about a sore throat?\\  
Person: Yes, I’ve had a sore throat.\\  
You: Have you experienced sneezing in the past 7 days, or not?\\  
Person: Yes, I have.\\  
You: How about muscle or body aches?\\  
Person: Yes, I’ve experienced that.\\  
You: Have you experienced headaches in the past 7 days, or not?\\  
Person: Yes, I have.\\  
You: Can you use a thermometer to take your temperature now?\\  
Person: Sure, just a moment. [Pause while Person takes their temperature] It’s 98.6°F.\\  
You: Thank you for providing that information. That concludes our survey. Thank you so much for your time and for your valuable responses. Have a great day!\\  
Person: You're welcome. Have a great day too!\\ 
  \end{spacing}

}

\section*{Persona generation procedure}
\label{app:persona_generation_description}

Fictitious persona generation is done in 2 main steps:
\begin{enumerate}
    \item Fictitious survey synthesis
    \item Fictitious persona synthesis
\end{enumerate}
To create a fictitious survey, we randomly select answers for each question according to one of three main strategies. The specific sampling strategy used for each question is shown in Table \ref{tab:sampling_strategies}.

\textbf{Strategy 1 (Uniform Random Selection).} We choose an answer uniformly at random from all available options. This strategy ensures that every response option is equally represented. We apply this method to Questions 3–5, which ask about race, household income, and highest degree obtained. Because these questions each have at least ten possible responses, using a uniform approach allows us to capture the full diversity of potential answers.

\textbf{Strategy 2 (Survey Response Probability-Based Selection).} We select answers based on probabilities derived from the frequency of responses in an actual COVID-19 impact survey conducted in April. After converting the observed frequencies to probabilities, we sample responses accordingly. This approach is used for Question 2 and Questions 12–33, which cover social conditions, mental health, physical health, and economic conditions. For Question 33, which deals with body temperature measurement, we first probabilistically determine whether the respondent refuses to provide a temperature. If not, we then select a temperature uniformly at random from the distribution of actual responses.

\textbf{Strategy 3 (Survey-Based Consistent Response Selection).} For Question 1 and Questions 6–11, we use actual respondent data from the April survey to maintain logical consistency among related questions. First, we randomly select a response to Question 6 (household size). We then filter the April survey dataset to include only those respondents who reported the same household size. From this filtered subset, we randomly choose one individual and use their reported numbers of household members in specific age categories, as well as their own age, to fill in our fictitious survey. This method ensures consistency between total household size and the number of people in each age group.
\begin{table}[h]  
    \centering  
    \caption{Response sampling strategy for each survey question.}  
    \begin{tabular}{ll}  
        \toprule  
        \textbf{Question} & \textbf{Sampling Strategy} \\  
        \midrule  
        1  & april survey-based consistent response                     \\  
        2  & survey response probability-based                \\  
        3  & uniform random                     \\  
        4  & uniform random                     \\  
        5  & uniform random                     \\  
        6  & april survey-based consistent response                     \\  
        7  & april survey-based consistent response                     \\  
        8  & april survey-based consistent response                     \\  
        9  & april survey-based consistent response                     \\  
        10 & april survey-based consistent response                     \\  
        11 & april survey-based consistent response                     \\  
        12 & april survey response probability-based               \\  
        13 & april survey response probability-based                \\  
        14 & april survey response probability-based                 \\  
        15 & april survey response probability-based                 \\  
        16 & april survey response probability-based                 \\  
        17 & april survey response probability-based                 \\  
        18 & april survey response probability-based                 \\  
        19 & april survey response probability-based                 \\  
        20 & april survey response probability-based                 \\  
        21 & april survey response probability-based                 \\  
        22 & april survey response probability-based                 \\  
        23 & april survey response probability-based                 \\  
        24 & april survey response probability-based                 \\  
        25 & april survey response probability-based                 \\  
        26 & april survey response probability-based                 \\  
        27 & april survey response probability-based                 \\  
        28 & april survey response probability-based                 \\  
        29 & april survey response probability-based                 \\  
        30 & april survey response probability-based                 \\  
        31 & april survey response probability-based                 \\  
        32 & april survey response probability-based                 \\  
        33 & april survey response probability-based        \\  
        \bottomrule  
    \end{tabular}  
    \label{tab:sampling_strategies}  
\end{table}  

\begin{figure}[h!]  
\centering  
\setlength{\fboxsep}{10pt}  
\fbox{  
    \parbox{0.95\textwidth}{ 
        {\fontfamily{pcr}\selectfont\small 
You are a diligent assistant tasked with analyzing survey conversation transcripts to deduce the respondent's answers to each of the 33 questions. Ensure that every question has a corresponding answer. Keep in mind the following guidelines:
\begin{enumerate}
    \item Deducing Answers: Responses to specific questions may sometimes be inferred from the broader conversation, even if the question was not explicitly asked. Use context to determine the best possible answer.
    \item Handling Transcription Errors: Conversations may contain transcription errors. Do your best to interpret the respondent's intended meaning accurately.
    \item Refusals or Skipped Questions: If the respondent explicitly refuses to answer or skips a question, label the answer as REFUSED. 
    Keep in mind that if the respondent explicitly refuses to answer one question, this may sometimes imply a refusal to answer other related questions as well.
    \item Ambiguity or Non-Matching Responses: If the respondent's answer does not align with any of the predefined response options or is ambiguous, label the answer as OTHER.
    \item Multiple Responses: If the respondent tells several matching response options within an answer to a question, consider the last response option they give as the final answer.
\end{enumerate}
Your goal is to carefully analyze the transcript and assign the most appropriate answer to each question based on the above criteria.

Below is the list of questions, each followed on the next line by its corresponding response options separated by semicolons:

[List of questions with response options follow here.]                     
}
    }  
}  
\caption{Assignment and role description for GPT-4o agent that analyzes conversation transcripts.}  
\label{fig:role_gpt4o_analyzer}  
\end{figure}

Once the fictitious survey is complete, we use a GPT-4o-based agent to generate a fictional persona. We start by defining the agent’s role and outlining the detailed procedure for creating the persona, as shown in Figure \ref{fig:role_persona_generator}. Then, using the survey data as input, we prompt the agent to produce the corresponding persona.
\begin{figure}[h!]  
\centering  
\setlength{\fboxsep}{10pt}  
\fbox{  
    \parbox{0.95\textwidth}{ 
        {\fontfamily{pcr}\selectfont\small 
You are an intelligent and resourceful assistant tasked with analyzing a survey consisting of 33 questions and their corresponding responses. 
Your objective is to craft a detailed, three-paragraph description of a fictional individual based on the survey answers you review. 
This description should be comprehensive enough that another person, upon reading it, would be able to accurately replicate the original responses to all 33 questions.
\\ \\
In the first paragraph, you will summarize and incorporate the answers to questions 1 through 11. 
The second paragraph will cover the responses provided for questions 12 through 22. 
Finally, the third paragraph will capture the answers to questions 23 through 33. 
The goal is to create a cohesive and vivid portrayal of the individual, ensuring the description reflects the essence of their survey answers in a clear and engaging manner.
\\ \\
Here are some additional guidelines to follow:
\begin{itemize}
     \item Make paragraphs as short as possible. No unnecessary information.
    \item When describing age, provide a specific number rather than a range.
    \item When describing household income, provide a specific figure that fits answer range. Don't state a range.
    \item When describing education attained, state the exact answer.
    \item When describing questions 14 though 18, state the answer in days.
    \item Make paragraphs as short as possible while maintaining narrative.
    \item Your description should not explicitly list or restate the answers to the survey questions. Exceptions for this rule are stated above. 
  Instead, craft a cohesive and natural narrative that captures the essence of the person's character and lifestyle.
\end{itemize}
}
    }  
}  
\caption{Assignment and role description for GPT-4o agent that generates persona from a fictitious survey.}  
\label{fig:role_persona_generator}  
\end{figure}

\section*{Instructions for analysis of transcripts}
\label{app:instruction_for_analyzer_gpt4o}

In our framework, GPT-4o infers answers to individual survey questions from conversation transcript. Initially, GPT-4o's role is describe with a prompt shown in Figure \ref{fig:role_gpt4o_analyzer}. Then GPT-4o is provided with full conversation and asked to produce responses to each of the questions. This process is repeated 5 times and the most frequent answer is chosen as the final response.

\section*{Post-study questionnaire}
\label{app:post_study_questionnaire}

After completing the surveys, participants were asked to evaluate their experience by indicating the extent to which they agreed or disagreed with the following statements:

\begin{enumerate}
\item The AI agent’s personality was realistic and engaging.
\item The AI agent appeared overly robotic.
\item The AI agent was welcoming and empathetic.
\item The AI agent seemed unfriendly.
\item The scope and purpose of the survey were clearly explained.
\item There was no clear indication of the survey’s purpose.
\item Surveys were conducted consistently and without errors.
\item The AI agent made frequent mistakes during the surveys.
\item The AI agent understood me well.
\item The AI agent failed to comprehend many of my responses.
\end{enumerate}

Participants rated each statement using a five-point Likert scale: strongly disagree, disagree, neither agree nor disagree, agree, and strongly agree. Additionally, they were invited to respond to an open-ended question: "What were the disadvantages of engaging with the AI agent, and what improvements could be made?"

\centering
\begin{table}[h!]  
\centering  
\begin{adjustbox}{width=\textwidth,center}
\renewcommand{\arraystretch}{1.2} % Adjust row height  
\setlength{\tabcolsep}{6pt} % Adjust column spacing  
\begin{tabular}{|p{1.11\textwidth}|} % Define fixed-width column  1
\hline  
 \textbf{Response} \\  
\hline  
 If I tell the AI agent that I am living on my own it does not need to loop through all questions about the age of the persons living in my household. Maybe just ask for the age of all persons and then put it in the age bins by itself. At some point it started to pronounce the end of each question weirdly by pitching the tone up. When I asked it to stop it stopped, but only for a few questions. \\  
\hline  
 The AI made some noticeable mistakes during the survey. For example, when asked to specify gender, I said "male," but it misinterpreted it as "mail." Additionally, when mentioning depression, it only acknowledged and showed empathy once across multiple attempts. The bot’s tone also felt plain and lacked enthusiasm or friendliness, making the experience less engaging. While it didn’t make many mistakes overall, the errors it did make were noticeable. It also struggled to handle ambiguous information and translate it into the appropriate survey response. I felt limited to giving short answers to the survey rather than giving natural/human responses. \\  
\hline  
 An AI agent could be more patient. For example, when I tried to say "fifteen thousand", the agent cut me off at "fifteen". \\  
\hline  
 The AI got hung up on particular words - female, "none" (said sarcastically/sardonically), and one other word. That was the most frustrating aspect of the interactions and also somewhat surprising. Sometimes I spoke over the AI and sometimes the AI spoke over what I was saying. Humans do the same thing, but I think the "turn taking" ability was not quite as good as the average human (however it was good enough it wasn't annoying, at least for me). One improvement might be to allow keypad entry of items (this would get around the problem with not being able to understand "none" -- I could just answer "0"). \\  
\hline  
 The AI had trouble with the words male and female. It sometimes struggled with numbers like "four". But overall it was actually better than most IVRs I have used. \\  
\hline  
 I probably would have hung up if it were a call out of the blue from an AI bot. That said, I thought the experience was good and would be intrigued if the first question was about whether I wanted to do the survey, and if I preferred talking to a human or to a realistic AI. There were a few things that the AI didn't quite understand, like if I said I lived alone or only with other adults, it still asked how many children in the different age brackets. Everything else was minor (like it seemed to want brief sentences such as "I'm a male" as opposed to just answering "male"). The rest I thought it did a good job, even when I answered in more natural language than just a yes or no. Lastly, it might help if at the outset it could be more explicit about what it can do, such as saying, "If at any point you need help or have any questions, simply ask and this AI can repond" \\  
\hline  
 Thanks for including me in your study! Pros: The AI Agent had a nice voice, the survey questions were clear and they understood me for the most part. Cons: The agent did not have a "personality" (is this what you were hoping for?) because there was no interaction that could have elicited this IMO. The AI Agent did not start with a warm greeting or ask me how I was doing, this seems like what a human would do. Also, it would have been good for them to give me an overview of why they were giving the survey i.e. "This will be a 5-7 minute survey, if you don't understand something please ask me to clarify." Maybe even say that it is an interactive AI agent and the intent is that you can converse with me. The AI agent did not understand my response of "Male" at least twice, the phrase "or not" sounded like a different voice, as did "you can take your temperature now" which was said in a juvenile or condescending tone. If the AI agent knew I lived alone, it should not have asked me the subsequent \# of persons in household questions (which were long). For the personas, I found that most said they were feeling lonely but not depressed, but when asked if depressed if they included that they were lonely I answered yes to depressed. If this was a human agent, they would have reiterated what I have said like, "you indicated you have felt lonely. . . to understand that more can you share if you've felt hopeless . . . depressed . . ." Side note: Regardless of if a survey is done by an AI Agent, human or self-service via an online tool, I don't particularly like to take surveys. If someone likes to take surveys and they prefer doing so with a human, I am not sure that the AI agent would be similar enough to a human to motivate them to take the survey. \\  
\hline  
 speak a bit slower for foreigners; skip some questions if the answer is none of the symptoms present. \\  
\hline  
\end{tabular} 
\end{adjustbox}
\caption{Responses to the post-study questionnaire question "What were the disadvantages of engaging with the AI agent, and what improvements could be made?" from 8 study participants.}  
\label{tab:poststudy_openended_responses}  
\end{table}

%TC:endignore 

\end{document}